# Extracting Symbolic Rules for Medical Diagnosis Problem

S. M. Kamruzzaman
Department of Computer Science and Engineering, Manarat International University, Bangladesh.
smzaman@gmail.com, smk.cse@manarat.ac.bd

**Abstract**
Neural networks (NNs) have been successfully applied to solve a variety of application problems involving classification and function approximation. Although backpropagation NNs generally predict better than decision trees do for pattern classification problems, they are often regarded as black boxes, i.e., their predictions cannot be explained as those of decision trees. In many applications, it is desirable to extract knowledge from trained NNs for the users to gain a better understanding of how the networks solve the problems. An algorithm is proposed and implemented to extract symbolic rules for medical diagnosis problem. Empirical study on three benchmarks classification problems, such as breast cancer, diabetes, and lenses demonstrates that the proposed algorithm generates high quality rules from NNs comparable with other methods in terms of number of rules, average number of conditions for a rule, and predictive accuracy.

## I. INTRODUCTION

The last two decades have seen a growing number of researchers and practitioners applying NNs for classification in a variety of real world applications. In some of these applications, it may be desirable to have a set of rules that explains the classification process of a trained network. The classification concept represented as rules is certainly more comprehensible to a human user than a collection of NNs weights.

While the predictive accuracy obtained by NNs is often higher than that of other methods or human experts, it is generally difficult to understand how the network arrives at a particular conclusion due to the complexity of the NNs architectures. It is often said that a NN is practically a "black box". Even for a network with only a single hidden layer, it is generally impossible to explain why a certain pattern is classified as a member of one class and another pattern as a member of another class.

This paper proposes a new rule extraction algorithm, called rule extraction from artificial neural networks (REANN) to extract symbolic rules from NNs for medical diagnosis problem. A standard three-layer feedforward NN is the basis of the algorithm. A four-phase training algorithm is proposed for backpropagation learning. In the first phase, the number of hidden nodes of the network is determined automatically in a constructive fashion by adding nodes one after another based on the performance of the network on training data. In the second phase, the NN is pruned such that irrelevant connections and input nodes are removed while its predictive accuracy is still maintained. In the third phase, the continuous activation values of the hidden nodes are discretized by using an efficient heuristic clustering algorithm. And finally in the fourth phase, rules are extracted by examining the discretized activation values of the hidden nodes using a rule extraction algorithm, REx.

## II. RELATED WORKS

There is quite a lot of literature on algorithms that extracts rules from trained NNs [1] [2]. Several approaches have been developed for extracting rules from a trained NN. Saito and Nakano [3] proposed a medical diagnosis expert system based on a multiplayer NN. They treated the network as black box and used it only to observe the effects on the network output caused by change the inputs.

Two methods for extracting rules from NN are described in Towell and Shavlik [4]. The first method is the subset algorithm [5], which searches for subsets of connections to a unit whose summed weight exceeds the bias of that node. The major problem with Subset algorithms is that the cost of finding all subsets grows as the size of the power set of the links to each node. The second method, the MofN algorithm [6], is an improvement of the Subset method that is designed to explicitly search for M-of-N rules from knowledge based NNs. Instead of considering a NN connection, groups of connections are checked for their contribution to a node's activation. This is done by clustering the NN connections.

H. Liu and S. T. Tan [7] proposes X2R, a simple and fast algorithm that can applied to both numeric and discrete data, and generate rules from datasets. It can generate perfect rules in the sense that the error rate of the rules is not worse than the inconsistency rate found in the original data. The rules generated by X2R, are order sensitive, i.e, the rules should be fired in sequence.

R. Setiono [8] proposes a rule extraction algorithm for extracting rules from pruned NNs for breast cancer diagnosis. The author describes how the activation values of a hidden node can be clustered such that only a finite and usually small number of discrete values need to be considered while at the same time maintaining the network accuracy. A small number of different discrete activation values and a small number of connections from the inputs to the hidden units will yield a set of compact rules for problem.

R. Setiono proposes a rule extraction algorithm named NeuroRule [9]. This algorithm extracts symbolic classification rule from a pruned network with a single hidden layer in two steps. First, rules that explain the network outputs are generated in terms of the discretized activation values of the hidden units. Second, rules that explain the discretized hidden unit activation values are generated in terms of the network inputs. When two sets of rules are merged, a DNF representation of network classification is obtained.

R. Setiono [10] proposes a rule extraction (RX) algorithm to extract rules from a pruned NN. The process of extracting rules from a trained NN can be made much easier if the complexity of the NN has first been removed. The pruning process attempts to eliminate as many connections as possible from the NN, while at the same time tries to maintain the prespecified accuracy rate.

R. Setiono, W. K. Leow and Jack M. Zurada [11] describes a method called rule extraction from function approximating neural networks (REFANN) for extracting rules from trained NNs for nonlinear regression. It is shown that REFANN produces rules that are almost as accurate as the original networks from which the rules are extracted.

### III. OBJECTIVE OF THE RESEARCH

Multilayer feedforward NNs trained by using the backpropagation-learning algorithm is limited to searching for a suitable set of weights in an apriori fixed network topology. Too small networks are unable to learn the problem well while overly large networks tend to overfit the training data, and consequently result in poor generalization performance. This paper proposes a hybrid approach with both constructive and pruning components for automatic determination of simplified NN architectures. The objective of the research are summarized as follows:
   i)  To develop an efficient algorithm for extracting symbolic rules from NNs for medical diagnosis problem to explain the functionality of NNs.
   ii) To find an efficient method for clustering the outputs of hidden nodes.
   iii) To extract concise rules with high predictive accuracy.

### IV. PROPOSED ALGORITHM

It is becoming increasingly apparent that without some form of explanation capability, the full potential of NNs may not be realized. The rapid and successful proliferation of applications incorporating NNs in many fields, such as commerce, science, industry, medicine etc., offers a clear testament to the capability of NN paradigm. Extracting symbolic rules from trained NN is one of the promising areas that are commonly used to explain the functionality of NNs. The aim of this section is to introduce a new algorithm to extract symbolic rules from trained NNs for medical diagnosis problem. The new algorithm is known as rule extraction from ANNs (REANN). Detailed descriptions of REANN are presented below.

#### A. THE *REANN* ALGORITHM

A standard three-layer feedforward NN is the basis of the proposed algorithm REANN. The hyperbolic tangent function, which can take any value in the interval [-1, 1], is used as the hidden node activation function. Rules are extracted from near optimal NN by using a new rule extraction algorithm, REx. The aim of REANN is to search for simple rules with high predictive accuracy. The major steps of REANN are summarized in Fig. 1 which are explained further as follows:

**Step 1** Create an initial NN architecture. The initial architecture has three layers, i.e. an input, an output, and a hidden layer. The number of nodes in the input and output layers is the same as the number of inputs and outputs of the problem. Initially, the hidden layer contains only one node. The number of nodes in the hidden layer is automatically determined by using a basic constructive algorithm. Randomly initialize all connection weights within a certain small range.

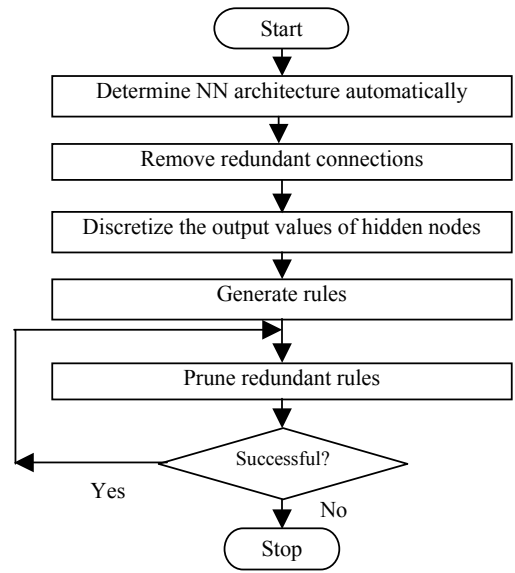

Fig. 1 Flow chart of the REANN algorithm.

**Step 2** Remove redundant input nodes, and connections between input nodes and hidden nodes and between hidden nodes and output nodes by using a basic pruning algorithm. When pruning is completed, the NN architecture contains only important nodes and connections. This architecture is saved for the next step.

**Step 3** Discretize the outputs of hidden nodes by using an efficient heuristic clustering algorithm. The reason for discretization is that the outputs of hidden nodes are continuous, thus rules are not readily extractable from the NN.

**Step 4** Generate rules that map the inputs and outputs relationships. The task of the rule generation is accomplished in three phases. In the first phase, rules are generated by using rule extraction algorithm, REx, to describe the outputs of NN in terms of the discretized output values of the hidden nodes. In the second phase, rules are generated by REx, to describe the discretized output values of the hidden nodes in terms of the inputs. Finally in the third phase, rules are generated by combining the rules generated in first and second phase.

**Step 5** Prune redundant rules generated in Step 4. Replace specific rules with more general ones.
**Step 6** Check the classification accuracy of the network. If the accuracy falls below an acceptable level, i.e. rule pruning is not successful then stop. Otherwise go to Step 5.

The rules extracted by REANN are compact and comprehensible, and do not involve any weight values. The accuracy of the rules from pruned networks is high as the accuracy of the original networks. The important features of REANN are the rule generated by REx is recursive in nature and is order insensitive, i.e, the rules need not be required to fire sequentially.

### B. RULE EXTRACTION ALGORITHM (REx)

Classification rules are sought in many areas from automatic knowledge acquisition [12] to data mining and NN rule extraction. This is because classification rules possess some attractive features. They are explicit, understandable and verifiable by domain experts, and can be modified, extended and passed on as modular knowledge. The steps of the Rule Extraction (REx) algorithm are summarized as follows:

**Step 1** Extract Rule:
  i=0; while (data is NOT empty/marked){
  generate Ri to cover the current pattern and differentiate it from patterns in other categories;
  remove/mark all patterns covered by Ri ; i++}
**Step 2** Cluster Rule:
  Cluster rules according to their class levels. Rules generated in Step 1 are grouped in terms of their class levels. In each rule cluster, redundant rules are eliminated; specific rules are replaced by more general rules.

**Step 3** Prune Rule:
  replace specific rules with more general ones;
  remove noise rules;
  eliminate redundant rules;
**Step 4** Check whether all patterns are covered by any rules. If yes then stop, otherwise continue.
**Step 5** Determine a default rule:
  A default rule is chosen when no rule can be applied to a pattern.

## V. EXPERIMENTAL STUDIES

This section evaluates the performance of REANN on three well-known benchmark classification problems. These are the breast cancer, diabetes, and lenses classification problems. They are widely used in machine learning and NN research. The data sets representing all the problems were real world data and obtained from the UCI machine learning benchmark repository.

### A. DATA SET DESCRIPTION

The following subsections briefly describe the data set used in this study. The characteristics of the data sets are summarized in Table I. The detailed descriptions of the data sets are available at ics.uci.edu in directory /pub/machine-learning-databases [13].

Table I Characteristics of data sets

| Data Sets | No. of Examples | Input Attributes | Output Classes |
|---|---|---|---|
| Breast Cancer | 699 | 9 | 2 |
| Diabetes | 768 | 8 | 2 |
| Lenses | 24 | 4 | 3 |

**A.1. The Breast Cancer Data**
The purpose of this problem was to diagnose a breast tumor as either benign or malignant based on cell descriptions gathered by microscopic examination. Input attributes were for instance the clump thickness, the uniformity of cell size and cell shape, the amount of marginal adhesion, and the frequency of bare nuclei.

**A.2. The Diabetes Data**
The objective of this data set was diagnosis of diabetes of Pima Indians. Based on personal data, such as age, number of times pregnant, and the results of medical examinations e.g. blood pressure, body mass index, result of glucose tolerance test, etc., try to decide whether a Pima Indian individual was diabetes positive or not.

**A.3 The Lenses Data**
The data set contains 24 examples and are complete and noise free. The examples highly simplified the problem. The attributes do not fully describe all the factors affecting the decision as to which type, if any, to fit.

### B. EXPERIMENTAL SETUP

In all experiments, one bias node with a fixed input 1 was used for hidden and output layers. The learning rate was set between [0.1, 1.0] and the weights were initialized to random values between [-1.0, 1.0]. Hyperbolic tangent function $\delta(y) = \frac{e^y - e^{-y}}{e^y + e^{-y}}$ is used as hidden node activation function and logistic sigmoid function $\sigma(y) = \frac{1}{1 + e^{-y}}$ as output node activation function.

In this study, all data sets representing the problems are divided into two sets. One is the training set and the other is the testing set. The numbers of examples in the training set and testing set are based on numbers in other works, in order to make comparison with those

works possible. The sizes of the training and testing data sets used in this study are given as follows:
*Breast cancer data set:* the first 350 examples are used for the training set and the rest 349 for the testing set.
*Diabetes data set:* the first 384 examples are used for the training set and the rest 384 for the testing set.

activation's were needed to maintain the accuracy of the network. The discrete values found by the heuristic clustering algorithm were 0.987, -0.986 and 0.004. Of the 350 training data, 238 patters have the first value, 106 have the second value and rest 6 patterns have third value.

Table II ANN architectures and training epochs for **breast cancer** data. The results were averaged over 10 independent runs.

|  | Initial Architecture | | Intermediate Architecture | | Final Architecture | | No. of Epoch |
|---|---|---|---|---|---|---|---|
|  | No. of Node | No. of Connection | No. of Node | No. of Connection | No. of Node | No. of Connection |  |
| Mean | 12 (9-1-2) | 11 | 12.7 | 18.1 | 6.8 | 5.8 | 233.2 |
| Min | 12 (9-1-2) | 11 | 12 | 11 | 5 | 5 | 222 |
| Max | 12 (9-1-2) | 11 | 14 | 33 | 10 | 9 | 245 |

Table III ANN architectures and training epochs for **diabetes** data. The results were averaged over 10 independent runs.

|  | Initial Architecture | | Intermediate Architecture | | Final Architecture | | No. of Epoch |
|---|---|---|---|---|---|---|---|
|  | No. of Node | No. of Connection | No. of Node | No. of Connection | No. of Node | No. of Connection |  |
| Mean | 11 (8-1-2) | 10 | 13.2 | 30 | 12.5 | 19.4 | 302.6 |
| Min | 11 (8-1-2) | 10 | 12 | 20 | 12 | 14 | 279 |
| Max | 11 (8-1-2) | 10 | 14 | 40 | 13 | 24 | 326 |

Table IV ANN architectures and training epochs for **lenses** data. The results were averaged over 10 independent runs.

|  | Initial Architecture | | Intermediate Architecture | | Final Architecture | | No. of Epoch |
|---|---|---|---|---|---|---|---|
|  | No. of Node | No. of Connection | No. of Node | No. of Connection | No. of Node | No. of Connection |  |
| Mean | 8 (4-1-3) | 7 | 9.1 | 14.7 | 8.9 | 12.1 | 109.2 |
| Min | 8 (4-1-3) | 7 | 8 | 7 | 8 | 7 | 97 |
| Max | 8 (4-1-3) | 7 | 10 | 21 | 10 | 17 | 128 |

**C. EXPERIMENTAL RESULTS**

Tables II-IV show NN architectures produced by REANN and training epochs over 10 independent runs on three benchmark classification problems.

The initial architecture was selected before applying the constructive algorithm, which was used to determine the number of nodes in the hidden layer. The intermediate architecture was the outcome of the constructive algorithm, and the final architecture was the outcome of pruning algorithm used in REANN.

It is seen that REANN can automatically determine compact ANN architectures. For example, for the breast cancer data, REANN produces more compact architecture. The average number of nodes and connections were 6.8 and 5.8 respectively; in most of the 10 runs 5 to 6 input nodes were pruned.

Fig. 2 shows the smallest of the pruned networks over 10 runs for breast cancer problem. The pruned network has only 1 hidden node and 5 connections. The accuracy of this network on the training data and testing data were 93.429% and 96.275% respectively. In this example only three input attributes $A_1$, $A_6$ and $A_9$ were important and only three discrete values of hidden node

The weight of the connection from the hidden node to the first output node was 3.0354 and to the second output node was –3.0354.

Figs. 3 shows the training time error for breast cancer problem. It was observed that the training error decreased and maintained almost constant for a long time after some training epochs and then fluctuates. The fluctuation was made due to the pruning process. As the network was retrained after completing the pruning process thus the training error again maintained almost constant value.

**C.1 EXTRACTED RULES**

The number of rules extracted by REANN and the accuracy of the rules in training and testing data sets were described in Table V. But the visualization of the rules in terms of the original attributes ware not discussed. The following subsections discussed the rules extracted by REANN in terms of the original attributes. The number of conditions per rule and the number of rules extracted were also visualized here.

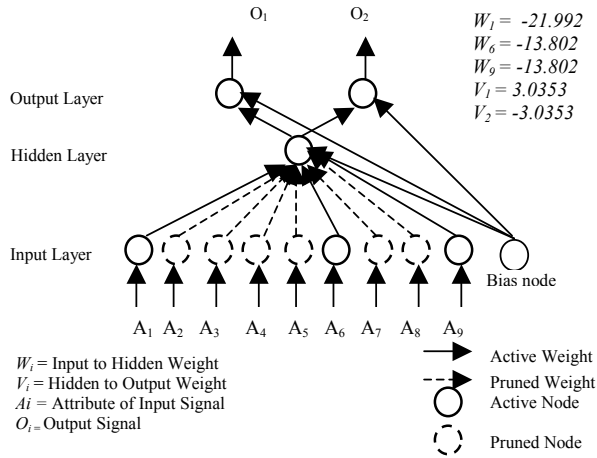

Fig. 2 A pruned network for breast cancer problem.

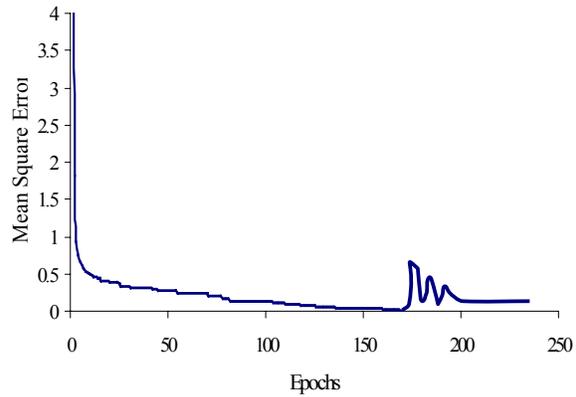

Fig. 3 Training time error for breast cancer problem.

### C.1.1 Breast Cancer Data

*Rule 1:* If Clump thickness ($A_1$) <= 0.6 and Bare nuclei ($A_6$) <= 0.5 and Mitosis($A_9$) <= 0.3, then benign

*Default Rule:* malignant.

### C.1.2 Diabetes Data

*Rule 1:* If Plasma glucose concentration ($A_2$) <= 0.64 and Age ($A_8$) <= 0.69 then tested negative

*Default Rule:* tested positive.

### C.1.3 Lenses Data

*Rule 1:* If Tear Production Rate ($A_4$) = reduce then no contact lenses
*Rule 2:* If Age ($A_1$) = presbyopic and Spectacle Prescription ($A_2$) = hypermetrope and Astigmatic ($A_3$) = yes then no contact lenses
*Rule 3:* If Age ($A_1$) = presbyopic and Spectacle Prescription ($A_2$) = myope and Astigmatic ($A_3$) = no then no contact lenses
*Rule 4:* If Age ($A_1$) = pre-presbyopic and Spectacle Prescription ($A_2$) = hypermetrope and Astigmatic ($A_3$) = yes and Tear Production Rate ($A_4$) = normal then no contact lenses
*Rule 5:* If Spectacle Prescription ($A_2$) = myope and Astigmatic ($A_3$) = yes and Tear Production Rate ($A_4$) = normal then hard contact lenses
*Rule 6:* If Age ($A_1$) = pre-presbyopic and Spectacle Prescription ($A_2$) = myope and Astigmatic ($A_3$) = yes and Tear Production Rate ($A_4$) = normal then hard contact lenses
*Rule 7:* If Age ($A_1$) = young and Spectacle Prescription ($A_2$) = myope and Astigmatic ($A_3$) = yes and Tear Production Rate ($A_4$) = normal then hard contact lenses

*Default Rule:* soft contact lenses.

Table V Number of extracted rules and rules accuracy for three benchmarks problems.

| Data Sets | No. of Extracted Rules | Rules Accuracy on Training Set | Rules Accuracy on Testing Set |
|---|---|---|---|
| Breast Cancer | 2 | 93.43 % | 96.28 % |
| Diabetes | 2 | 72.14 % | 76.56 % |
| Lenses | 8 | 100 % | 100 % |

Table V shows number of extracted rules and rules accuracy for three benchmark problems. It was observed that two rules were sufficient to solve breast cancer and diabetes problems. The accuracy was 100% for lenses classification. This data sets having lower number of examples.

### VI. COMPARISON

This section compares experimental results of REANN with the results of other works. The primary aim of this work is not to exhaustively compare REANN with all other works, but to evaluate REANN in order to gain a deeper understanding of rule extraction.

Table VI compares REANN results of breast cancer problem with those produced by NN RULES [9], DT RULES [9], C4.5 [12], NN-C4.5 [14], OC1 [14], and CART [15] algorithms. REANN achieved best performance although NN RULES was closest second. But number of rules extracted by REANN are 2 whereas these were 4 for NN RULES.

Table VII compares REANN results of diabetes data with those produced by NN RULES, C4.5, NN-C4.5, OC1, and CART algorithms.

REANN achieved 76.56% accuracy although NN-C4.5 was closest second with 76.4% accuracy. Due to the high noise level, the diabetes problem is one of the most challenging problems in our experiments. REANN has outperformed all other algorithms.

Table VI Performance comparison of REANN with other algorithms for **breast cancer** data

| Data Set | Feature | REANN | NN RULES | DT RULES | C4.5 | NN-C4.5 | OC1 | CART |
|---|---|---|---|---|---|---|---|---|
| Breast Cancer | No. of Rules | 2 | 4 | 7 | - | - | - | - |
| | Avg. No. of Conditions | 3 | 3 | 1.75 | | | | |
| | Accuracy % | 96.28 | 96 | 95.5 | 95.3 | 96.1 | 94.99 | 94.71 |

Table VII Performance comparison of REANN with other algorithms for **diabetes** data

| Data Set | Feature | REANN | NN RULES | C4.5 | NN-C4.5 | OC1 | CART |
|---|---|---|---|---|---|---|---|
| Diabetes | No. of Rules | 2 | 4 | | | | |
| | Avg. No. of Conditions | 2 | 3 | | | | |
| | Accuracy % | 76.56 | 76.32 | 70.9 | 76.4 | 72.4 | 72.4 |

Table VIII Performance comparison of REANN with other algorithm for **lenses** data.

| Data set | Feature | REANN | PRISM |
|---|---|---|---|
| Lenses | No. of Rules | 8 | 9 |
| | Avg. No. of Conditions | 3 | - |
| | Accuracy % | 100.0 | 100.0 |

Table VIII compares REANN results of lenses data with those produced by PRISM [16]. Both algorithms achieved 100% accuracy because the lower number of examples. Number of extracted rules by REANN are 8 whereas these were 9 for PRISM.

## VII. CONCLUSIONS

NNs are often viewed as black boxes. While their predictive accuracy is high, one usually cannot understand why a particular outcome is predicted due to the complexity of the network. This work is an attempted to open up these black boxes by extracting symbolic rules from it through the proposed efficient rule extraction algorithm REANN for medical diagnosis problem.

The REANN algorithm can extract concise rules from standard feedforward NN. Network training and pruning is done via the simple and widely used backpropagation method. No restriction is imposed on the activation values of hidden nodes or output nodes. An important feature of rule extraction algorithm, REx, is its recursive nature. They are concise, comprehensible, order insensitive and do not involve any weight values. The accuracy of the rules from a pruned network is as high as the accuracy of the fully connected network.

Extensive experiments have been carried out in this study to evaluate how well REANN performed on three benchmark classification problems in NNs including breast cancer, diabetes, and lenses in comparison with other algorithms. In almost all cases, REANN outperformed the others. With the rules extracted by the method introduced here, NNs should no longer be regarded as black boxes.